# Seizure Prediction using Bidirectional LSTM

Hazrat Ali[1[0000-0003-3058-5794]], Feroz Karim [2], Junaid Javed Qureshi[1], Adnan Omer Abu-assba[3] and Mohammad Farhad Bulbul[4]

[1] Department of Electrical and Computer Engineering,
COMSATS University Islamabad, Abbottabad Campus, Abbottabad, Pakistan
[2] Institute for Interdisciplinary Information Sciences,
Tsinghua University, Beijing, China.
3 Arab Open University-Palestine, Ramallah, Palestine.
[4] Department of Mathematics, Jashore University of Science and Technology, Bangladesh.
hazratali@cuiatd.edu.pk, ferozkundan@gmail.com,
jjqureshi123@gmail.com, farhad@just.edu.bd

**Abstract.** Approximately, 50 million people in the world are affected by epilepsy. For patients, the anti-epileptic drugs are not always useful and these drugs may have undesired side effects on a patient's health. If the seizure is predicted the patients will have enough time to take preventive measures. The purpose of this work is to investigate the application of bidirectional LSTM for seizure prediction. In this paper, we trained EEG data from canines on a double Bidirectional LSTM layer followed by a fully connected layer. The data was provided in the form of a Kaggle competition by American Epilepsy Society. The main task was to classify the interictal and preictal EEG clips. Using this model, we obtained an AUC of 0.84 on the test dataset. Which shows that our classifier's performance is above chance level on unseen data. The comparison with the previous work shows that the use of bidirectional LSTM networks can achieve significantly better results than SVM and GRU networks.

**Keywords:** Seizure Prediction, Bidirectional LSTM, Deep Learning, EEG

## 1 Introduction

Epilepsy is a neurological disorder characterized by spontaneous seizures. Approximately, 50 million people in the world are affected by epilepsy and roughly 80% of them belong to low- and middle-income countries [1]." For 20-40% of patients with epilepsy, medications are not effective – and even after surgical removal of epilepsy-causing brain tissue; many patients continue to experience spontaneous seizures. Although seizures occur infrequently, patients with epilepsy experience persistent anxiety due to the possibility of a seizure occurring" [2]. Similarly, the care taker person also suffers due to uncertain conditions of the patient. According to multi-center clinical studies, 6.2% of patients reported premonitory symptoms [3], and some of the epilepsy patients interviewed felt "auras" [4]. All these indicated that seizures might be predicted. Early detection can enable a patient as well as the care-taker to ensure precautionary steps for minimizing the associated risks by bringing the patient into a more comfortable and safer environment or bring him/her to rest if moving.
The brain activity can be classified into four states: Interictal (between seizures, or baseline), Preictal (prior to seizure), Ictal (seizure), and Post-ictal (after seizures) [2]. The *American Epilepsy Society* (AES) announced a competition on the platform of Kaggle (Kaggle, Inc) on the seizure prediction task. Our task is to develop a model to differentiate between Preictal and Interictal states. The data is collected from seven subjects, five canines, and two humans. The data consisted of 10-minute clips labeled "Preictal" for pre-seizure data segments, or "Interictal" for non-seizure data segments. The participants of the competition are required to distinguish between interictal and preictal clips. Preictal data segments are the six 10-minutes clips prior to seizure with a 5-minute margin from seizure, as shown in Figure 1. This 5-minute interval has been left to predict seizure on a minimum onset of 5 minutes so that the patient may be enabled to take preventive measures before occurrence of seizure. From figure 1, it can also be seen that there are more than one recordings for each segment in the figure. These recordings are from different electrodes placed at different positions of the brain.

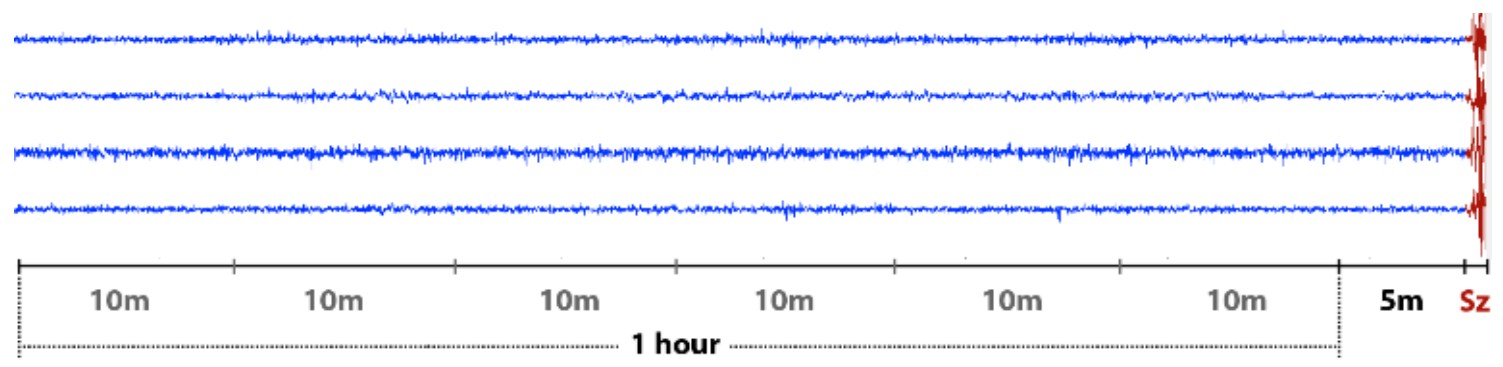

**Fig. 1.** Ten minutes pre-ictal clips and a five minute interval before the seizure can be seen (Source: kaggle.com)

The classification of data into preictal and interictal segments makes this a binary classification task, in which the computational model has to predict the class of a given clip. The evaluation method used in the competition is area under the ROC curve (AUC). A higher AUC means that the model has given a higher probability to preictal clips. For a perfect classification result, the AUC would be 1. The computational model is trained on the intracranial encephalogram (iEEG) data from different subjects and tested on the unseen data.
In this paper, we investigate the use of bidirectional long short-term memory (LSTM) network for seizure prediction, which to the best of our knowledge has not been addressed before. In a recent study, Generalized Linear Models (GLM), Support Vector Machines (SVM), Random Forests and Convolutional Neural Networks (CNN) were trained and achieved AUC score in the range of 0.82 – 0.86 range [5].
CNN models have been more successful on image data or where a problem can be addressed such that the data is treated as images (e.g., spectrogram of speech data). However, for time-series data, RNN model would be a better suit. In a comparative study of RNN and CNN for natural language processing, which mainly involves processing of sequential data, it can be proved that RNN outperforms CNN on most of the tasks [6]. CNN model would not be able to capture the time depenencies of the data while an LSTM model look after the time dependencies both in forward and backward direction. This is one major motivation for the use of Bi-directional LSTM on the given EEG data. The rest of the paper is organized as follows: In Section 2, we provide a description of the data and explain the features. Section 3 discusses the LSTM model usage. Results are reported in Section 4. Finally, the paper is concluded in Section 5.

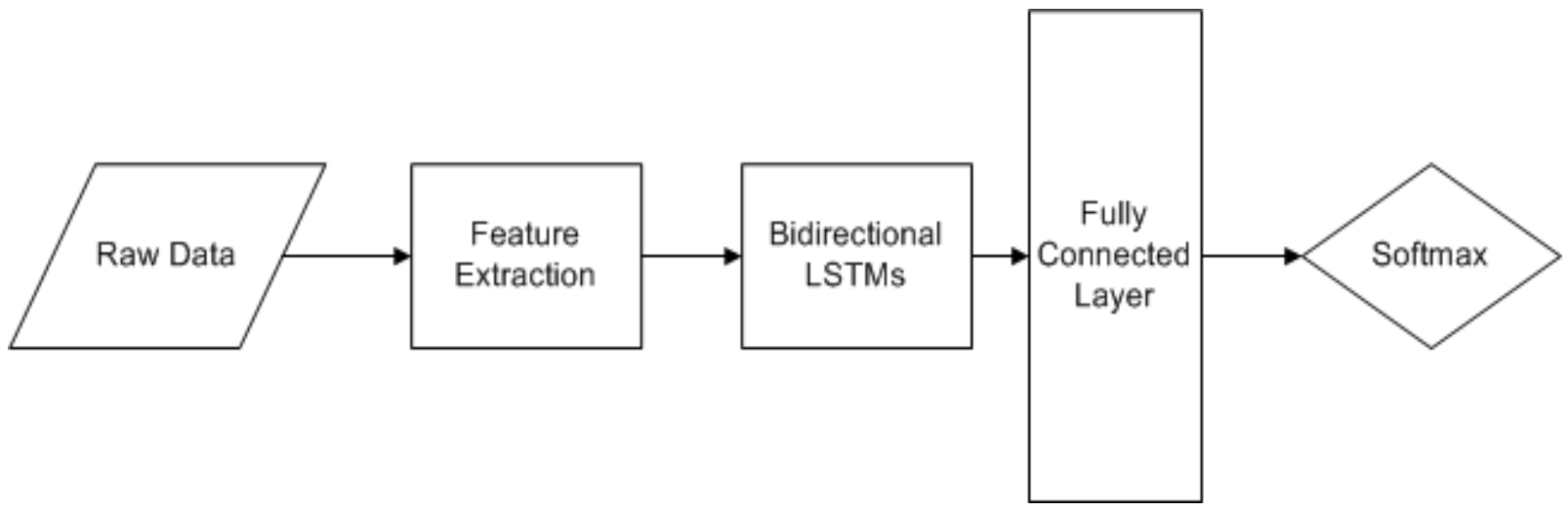

**Fig. 2.** Diagram showing the overall approach used in this work

**Table 1.** Total number of labeled clips available in the dataset

| Subject | Total Clips | Preictal Clips | Interictal Clips |
|---|---|---|---|
| **Dog1** | 504 | 24 | 480 |
| **Dog2** | 542 | 42 | 500 |
| **Dog3** | 1512 | 72 | 1440 |
| **Dog4** | 901 | 97 | 804 |

## 2 Data Description and Feature Extraction

The iEEG Data used for this paper was recorded as 10-minute long clips from different subjects with a sampling frequency of 400 Hz. The data is provided by AES and hosted on Kaggle. In this work, data from four canines has been used for training of the model [7], [8].

Each clip consists of a 10-minute long recording from 16 different electrodes at a sampling frequency of 400 Hz. This implies that each clip has 600 sec × 400 Hz ×16 channels =3.84 million samples. Table 1 shows that canines we selected to have 3459 recordings. This is a huge number. Thus, in order to train the model, it is important to extract useful features.

Feature extraction is done to reduce the dimensionality and extract fruitful information from the data. For this purpose, each 10-minute clip is split into 20 smaller clips of 30 seconds each. As each clip consists of 16 channels recordings so at the end of this process, we get 20 × 16=320 segments from each clip. Features extracted from the data are stated below.

### 2.1 Power Spectral Intensity

Power spectral intensity (PSI) is computed for each bin using PyEEG library [9]. PyEEG follows the below pattern to compute PSI.

i) Fast Fourier Transform, X= [$X_1$, $X_2$, $X_3$……., $X_N$] is obtained for each 30-second clip.

ii) PSI is calculated in eight different frequency bins using this mathematical relation.

$$PSI = \sum_{i=f_1}^{f_2} |X_i|,$$

Where $f_1$ and $f_2$ are the lower and higher values of a bin. The bins are: {[0.1, 4], [4, 8], [8, 12], [12, 30], [30, 50], [50, 70], [70, 100], [100, 180]} in Hz. The first four frequency bins correspond approximately to the δ, θ, α and β bands respectively. These bands are used frequently in neurophysiology [10].

### 2.2 Standard Deviation

In addition, the standard deviation for each 30-second segment is measured as one of the features. For a smaller clip, 9 features are extracted from each channel resulting in a total of 16 × 9=144 features obtained for that clip. From a single 10-minute clip, 2880 features are mined. A random shuffle is applied to the data for faster convergence [11]. After this, the data is divided into training and test sets. A total of 2900 samples are used to train the model, and the rests 559 are used as a test set.

## 3 Model Training

We train Bidirectional LSTM units followed by a fully connected layer of artificial neural network (ANN). Figure 2 shows the overall approach used in this work. The features extracted from the data are used to train the model. At the last layer, a softmax layer is used to classify pre-ictal and inter-ictal EEG clips.

In this work, we have considered only the Bidirectional LSTM model, and the model is trained without any regularization technique applied. In this section, we will discuss Bidirectional LSTM [12] unit and how we used these to obtain good results.

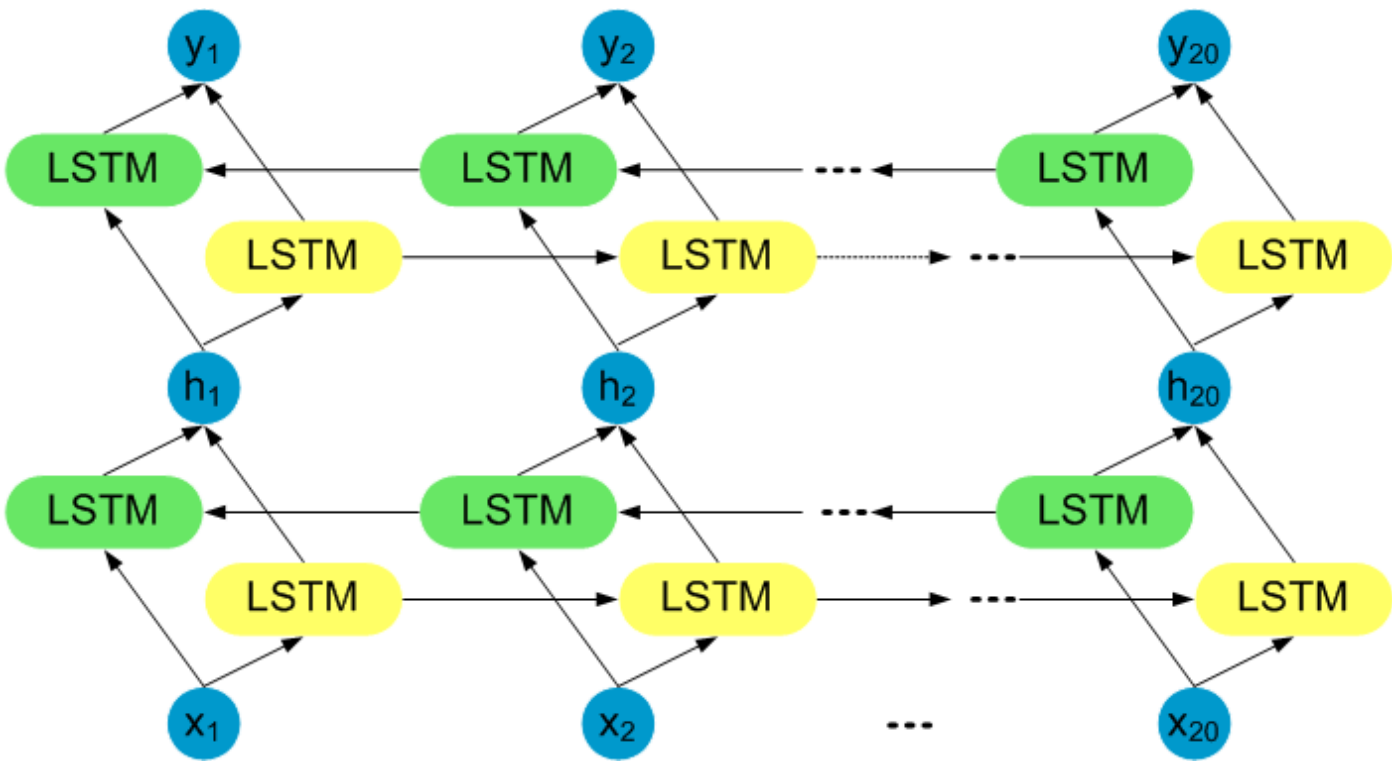

**Fig. 3.** A double Bidirectional LSTM layer network: each layer contains 512 hidden nodes

LSTM unit is a variant of Recurrent Neural Networks (RNN). Due to the vanishing and exploding gradient problems, it is hard to train standard RNN [13], [14]. In an LSTM, the activation function is an identity function having derivative equal to 1. This stops the gradient from vanishing or exploding and rather keeps it constant. The architecture of the LSTM was presented in [11] and is formulated as:

$$f_t = \sigma_g(W_f x_t + U_f h_{t-1} + b_f) \quad (1)$$
$$i_t = \sigma_g(W_i x_t + U_i h_{t-1} + b_i) \quad (2)$$
$$o_t = \sigma_g(W_o x_t + U_o h_{t-1} + b_o) \quad (3)$$
$$\tilde{c}_t = tanh(W_c x_t + U_c h_{t-1} + b_c) \quad (4)$$

where $W_f$, $W_i$, $W_o$, and $W_C$ are the weight matrices mapping the hidden layer input to the three gates and the input cell state, while the $U_f$, $U_i$ , $U_o$, and $U_C$ are the weight matrices connecting the previous cell output state to the three gates and the input cell state. The $b_f$, $b_i$, $b_o$, and $b_C$ are four bias vectors. The $\sigma_g$ is the gate activation function, which normally is the sigmoid function, and the *tanh* is the hyperbolic tangent function. Based on the results of the above four equations, at each time iteration $t$, the cell output state, $C_t$, and the layer output, $h_t$, can be calculated as follows:

$$C_t = f_t \times C_{t-1} + i_t \times \tilde{C}_t \quad (5)$$
$$h_t = o_t \times \tan(C_t) \quad (6)$$

The LSTM architecture selected for this problem in Figure 3 consisted of 20 forward and 20 backward LSTM cells per layer. Each cell accepts 144-dimensional vector corresponding to a single 30-second clip. The output is obtained by connecting the final LSTM cell output to fully connected layer that outputs the probabilities for both classes. We used Xavier initialization method [15] to initialize our fully connected layer variables. Adam optimization algorithm [16] is used for the training of network with a batch size of 290. A learning rate of $10^{-3}$ is used for training.

# 4 Results

The performance of the model is tested with Area under the ROC curve (AUC). The importance is given to correctly predict pre-ictal events. Hence, the goal is to maximize true positive rate (TPR) and to keep a false-positive rate (FPR) reasonably low. And thus, we use Receiver Operating Characteristics (ROC) curve and Area Under ROC curve (AUC) to evaluate the performance of the model. ROC curve is a plot between TPR and FPR. The higher values of AUC indicate the better results and vice versa.

**Table 2.** Maximum AUC achieved using Bidirectional LSTM

| | **Training Dataset AUC** | **Test Dataset AUC** |
|---|---|---|
| **Split1** | 0.88 | 0.84 |
| **Split2** | 0.93 | 0.81 |

**Table 3.** AUC obtained by [17] using 2 layer GRU network

| | **Validation dataset AUC** | **Test set AUC** |
|---|---|---|
| **Split 1** | 0.94 | 0.46 |
| **Split 2** | 0.69 | 0.61 |
| **Split 3** | 0.87 | 0.63 |
| **Split 4** | 0.82 | 0.64 |
| **Split 5** | 0.86 | 0.71 |

**Table 4.** Average AUC obtained by [18] using three different classifiers.

| Model | Average AUC |
|---|---|
| Linear Least Squares | 0.78 |
| Linear Discriminant Analysis | 0.69 |
| Regularized SVM | 0.80 |

To calculate TPR and FPR, we need True Positives (TP), False Positives (FP), False Negative (FN), and True Negative (TN). TP is the number of examples predicted pre-ictal and labeled pre-ictal as well. FP is the number of examples incorrectly predicted as pre-ictal. FN is the number of examples predicted incorrectly as inter-ictal. TN is defined as the number of examples predicted correctly inter-ictal by the classifier.

The results obtained using the double Bidirectional LSTM layers are encouraging. Two different splits of Training and Test datasets are used in this work. The AUC obtained in Split1 and Split2 is 0.84 and 0.81 respectively, which are better than compared to the AUC obtained by [17]. The maximum AUC reported by [17] is 0.71 and the maximum AUC reported by [18] is 0.80. This shows that our approach achieves better AUC. Table 2, Table 3 and Table 4 show comparison of results for proposed method with those reported in [17] and [18].

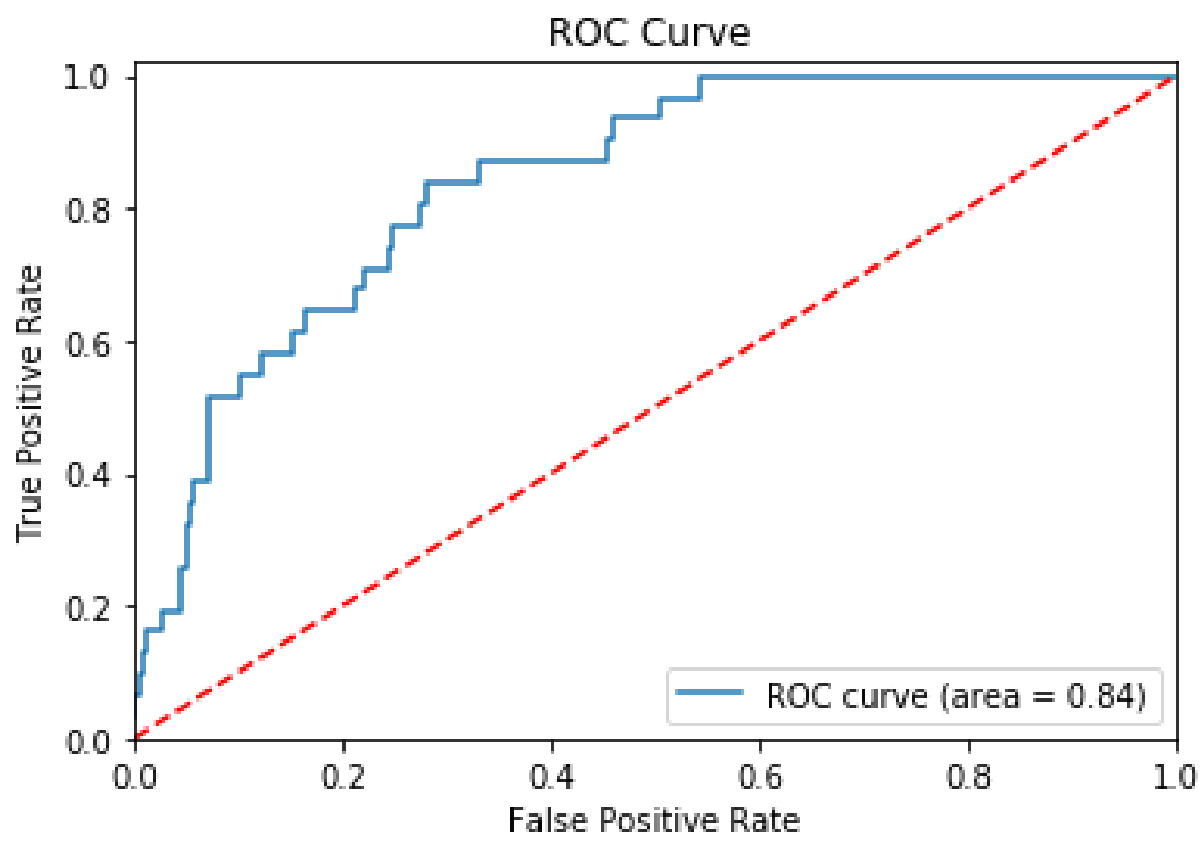

**Fig. 4.** Receiver operator characteristics for the split1 test dataset

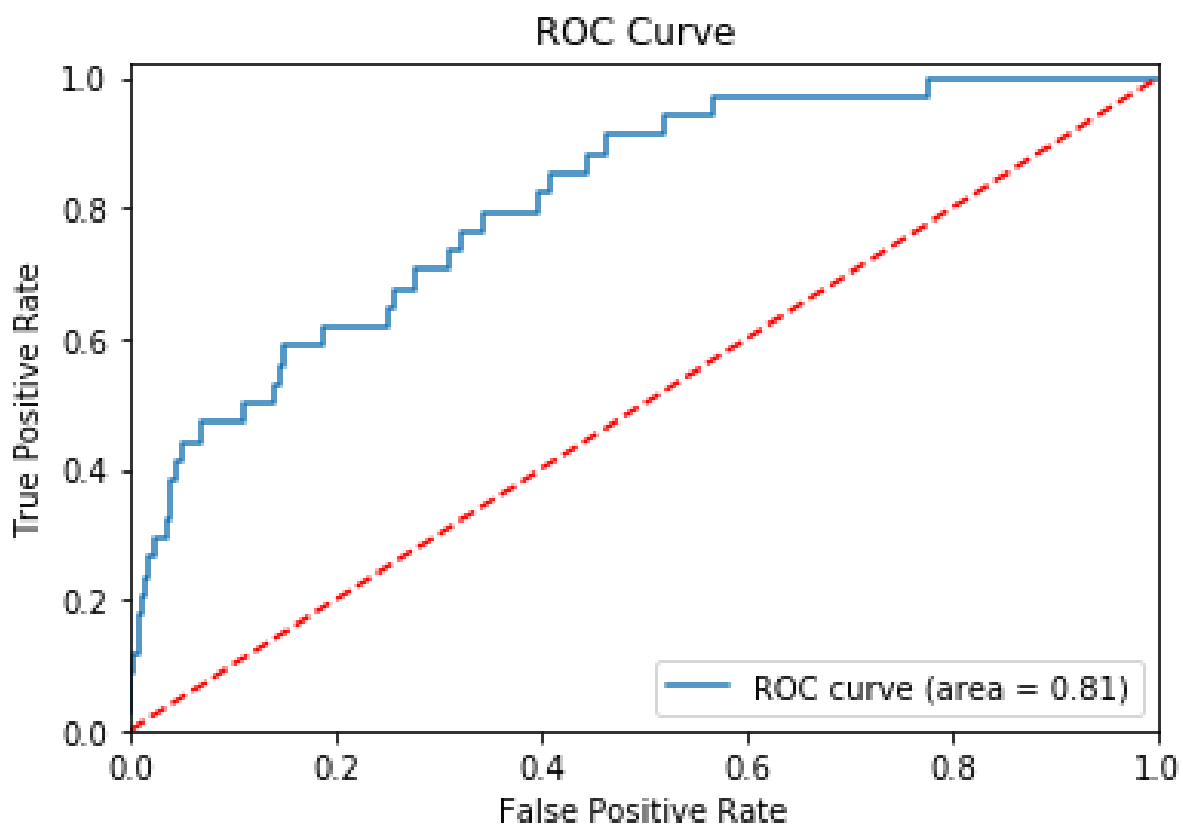

**Fig. 5.** Receiver operator characteristics for the split2 test dataset

The ROC curves for both split1 and split2 are shown in Figure 4 and Figure 5 respectively. From these figures, we can clearly see that the ratio of true positives is higher than false positives. The similarity between the AUC obtained for training and test sets indicates that we do not have an overfitting situation. To compare our results, we selected [17] as the authors have used the same subjects (data) with similar features' sets and trained a GRU model. In comparison, we have used a bidirectional LSTM which has given us better performance.

## 5 Conclusion and future work

In this paper, we have investigated the use of Bidirectional LSTM on EEG data for seizure prediction. A double Bidirectional LSTM layer was trained on the features extracted from raw data. The proposed model has shown promising results when tested on unseen test set. We received test set AUC of 0.84 and 0.81 for Split1 and Split2 respectively, which is better than AUC values 0.71 and 0.80, reported by [17] and [18] respectively. The predictions of the model with unseen data are notable. We did not allow the model to overfit and the test set performance is in close resemblance with performance for train set. The EEG data is typically challenging for understanding by humans, but with machine learning tools, we can process it and extract useful features from it. In future, much longer recordings of EEG data can be used to train the model which may then help to have even better insights into the data.

## References

1. World Health Organization. Atlas: epilepsy care in the world. Geneva: World Health Organization; 2005
2. AES Seizure Prediction Challenge,“ https://www.kaggle.com/c/seizure-prediction,” 2014.
3. A. Schulze-Bonhage, C. Kurth, A. Carius, B. J. Steinhoff, and T. Mayer, “Seizure anticipation by patients with focal and generalized epilepsy: a multicentre assessment of premonitory symptoms,” Epilepsy research, vol. 70, no. 1, pp. 83–88, 2006.
4. P. Rajna, B. Clemens, E. Csibri, E. Dobos, A. Geregely, M. Gottschal, I. Gyorgy, A. Horvath, F. Horvath, L. Mez ¨ ofi¨ et al., “Hungarian multicentre epidemiologic study of the warning and initial symptoms (prodrome, aura) of epileptic seizures,” Seizure, vol. 6, no. 5, pp. 361– 368, 1997.
5. Benjamin H. Brinkmann, Joost Wagenaar, Drew Abbot, Phillip Adkins, Simone C. Bosshard, Min Chen, Quang M. Tieng, Jialune He, F. J. Munoz-Almaraz, Paloma Botella- ˜ Rocamora, Juan Pardo, Francisco Zamora-Martinez, Michael Hills, Wei Wu, Iryna Korshunova, Will Cukierski, Charles Vite, Edward E. Patterson, Brian Litt, and Gregory A. Worrell. Crowdsourcing reproducible seizure forecasting in human and canine epilepsy. Brain, 2016.
6. Wenpeng Yin, Katharina Kann, Mo Yu, and Hinrich Schütze “Comparative Study of CNN and RNN for Natural Language Processing,” arXiv:1702.01923v1 [cs.CL], Feb. 2017.
7. H. Potschka, A. Fischer, E.-L. R¨uden, V. H¨ulsmeyer, and W. Baumg¨artner, “Canine epilepsy as a translational model?” Epilepsia 54, 571 (2013).
8. E. E. Patterson, “Canine epilepsy: an underutilized model,” ILAR Journal 55, 182 (2014).
9. Forrest Sheng Bao, Xin Liu, and Christina Zhang, “PyEEG: An Open Source Python Module for EEG/MEG Feature Extraction,” Computational Intelligence and Neuroscience, vol. 2011, Article ID 406391, 7 pages, 2011.
10. G. Deuschl, A. Eisen, et al., “Recommendations for the practice of clinical neurophysiology: guidelines of the international federation of clinical neurophysiology,” (1999).
11. Q. Meng, W. Chen, Y. Wang, Z.-M. Ma, and T.-Y. Liu, “Convergence analysis of distributed stochastic gradient descent with shuffling,” arXiv preprint arXiv:1709.10432, 2017.
12. Z. Cui, Ruimin Ke, Y. Wang, “Deep Bidirectional and Unidirectional LSTM Recurrent Neural Network for Network-wide Traffic Speed Prediction”, arXiv:1801.02143v1 [cs.LG], Jan. 2018.

13. Y. Bengio, P. Simard, and P. Frasconi, “Learning Long-Term Dependencies with Gradient Descent is Difficult,” IEEE Trans. Neural Networks, vol. 5, no. 2, pp. 157–166, 1994..
14. R. Pascanu, T. Mikolov, and Y. Bengio, “On the difficulty of training recurrent neural networks,” in International Conference on Machine Learning, 2013, no. 2, pp. 1310–1318.
15. Glorot, X.; Bengio, Y. “Understanding the difficulty of training deep feedforward neural networks”. Proceedings of the Thirteenth International Conference on Artificial Intelligence and Statistics, in PMLR, 9:249-256, 2010
16. D. Kingma and J. Ba, “Adam: A method for stochastic optimization,” arXiv preprint arXiv:1412.6980 (2014).
17. M. Larmuseau, ‘Epileptic seizure prediction using deep learning’, MS Thesis, Ghent University, 2016.
18. M. Gonzenbach, ‘Prediction of epileptic seizures using EEG data’, MS Thesis, ETH Zurich, 2015.